%% file: iclr2019_conference.tex
\documentclass{article} 
\usepackage{iclr2019_conference,times}

\input{math_commands.tex}

\usepackage{hyperref}
\usepackage{url}
\usepackage{graphicx}  
\newcommand{\norm}[1]{\left\lVert#1\right\rVert}
\graphicspath{{images/}}

\title{Towards Reducing Bias in Gender Classification}

\author{Komal K. Teru \& Aishik Chakraborty \\School of Computer Science\\
McGill University\\
Montreal, Quebec, Canada \\
\texttt{\{komal.teru, aishik.chakraborty\}@mail.mcgill.ca} \\
}

\iclrfinalcopy 
\begin{document}

\maketitle

\begin{abstract}
Societal bias towards certain communities is a big problem that affects a lot of machine learning systems. This work aims at addressing the racial bias present in many modern gender recognition systems. We learn race invariant representations of human faces with an adversarially trained autoencoder model. We show that such representations help us achieve less biased performance in gender classification. We use variance in classification accuracy across different races as a surrogate for racial bias of the model and achieve a drop of over 40\% in variance with race invariant representations.

\end{abstract}

\section{Introduction}
AI systems are being currently used in many applications in our society, starting from adding filters to make our face look good on social media to making recruiting decisions. Standalone AI systems are not used in tasks such as determining how long someone stays in prison, rather they are used in pipeline. For example, facial recognition systems are typically used to identify suspects, even though, they are finally not used to determine who committed a crime. Even if an AI system is used in the pipeline, one wrong decision can have huge impact in the life of a person. Thus, many people ask the question, 'Are AI systems biased in making their choices or are they fair to everyone?'. 

The question of fairness is scientifically difficult to  define. However, that does not stop people from delving into these topics and asking the hard questions. It is known that many AI systems, like tools which do facial recognition rely on machine learning algorithms that are trained with labeled data. The datasets available in the wild can be poorly constructed and can contain various societal biases. It has recently come to light that ML algorithms trained with biased data can result in learning discriminating features which are biased towards certain communities. In other words, these ML algorithms learn the societal biases that we want to avoid. \cite{bolukbasi2016man}  showed  that  popular word embedding algorithms like Word2Vec(~\cite{mikolov2013efficient}) encodes societal gender biases. The authors used Word2Vec on the word-analogy task and found that  man  is  to computer programmer as woman is to "?" was completed with “homemaker”. This is the kind of bias that we ideally want our models to be free of. These embeddings are very popular and people unaware of these biases use these embeddings in their models, thereby including such stereotypes in their models as well. 

In this work, we therefore take steps towards reducing bias in the domain of gender classification. A recent work by \cite{buolamwini2018gender} has found that facial recognition and gender classification systems are biased towards people having lighter skin tone. We therefore, investigate this problem and try to reduce such biases in ML systems. We do so by learning race invariant representations of an image of a face for which we wish to classify gender, using a kind of adversarial loss proposed in the paper \cite{lample2017fader}. We use such race-invariant representations to do gender classification in the hope that our gender classifier will perform well for all race categories, instead of being biased towards one race category.

\section{Related Work}
\textbf{Learning Invariant Representations.} The authors in \cite{lample2017fader} propose Fader Networks that can be used to learn attribute invariant representations. The model consists of a discriminator that tries to predict input attribute $c$, given a latent representation of the input image $z$, and is trained by an adversarial loss function that tries to fool the discriminator. This work is also related to the work on learning invariant latent spaces using adversarial training in domain adaptation \cite{ganin2016domain}. Similar work include the domain of fair classification like \cite{edwards2015censoring}, where we want to remove sensitive information from images. The concept used is also similar to robust inference \cite{louppe2017learning}. This kind of work has been leveraged in the domain of Unsupervised Machine Translation like ~\cite{lample2017unsupervised}, where we can think of the "attribute" under consideration as the kind of language we are considering, and thus, there the latent representation $z$ can be thought of like a kind of inter-lingua. A similar idea has also been leveraged by ~\cite{conneau2017word} in the cross-lingual word translation task without using parallel data.

\textbf{Bias in Machine Learning.} \cite{bolukbasi2016man} showed that due to the quality of data that we train our models on, popular word embeddings like Word2Vec(\cite{mikolov2013efficient}) exhibit shocking societal gender biases. Similarly, there are other works in natural language processing systems on bias like sentiment analysis(\cite{kiritchenko2018examining}) and semantic role labelling(\cite{yatskar2017commonly}). There has been work in NLP on reducing gender bias like \cite{zhao2017men}. However, in the field of computer vision, there does not seem to be a lot of work on bias in facial recognition systems. \cite{klare2012face} show that some face recognition systems misidentify people of color, young people and also women at high rates. The paper by ~\cite{buolamwini2018gender} focuses on the fact that state of the art facial recognition systems does well on people with lighter skin tone.

\section{Model}

As mentioned earlier, we use adversarially trained autoencoder model to obtain attribute (in our case, race) invariant representations of input images. In the following sections we discuss the architecture and implementation in detail. Next, we discuss the models used for gender classification and approaches followed to make a fair comparison of different models.

\subsection{Fader Net Architecture}

\begin{figure}[h!]
\includegraphics[width=\linewidth]{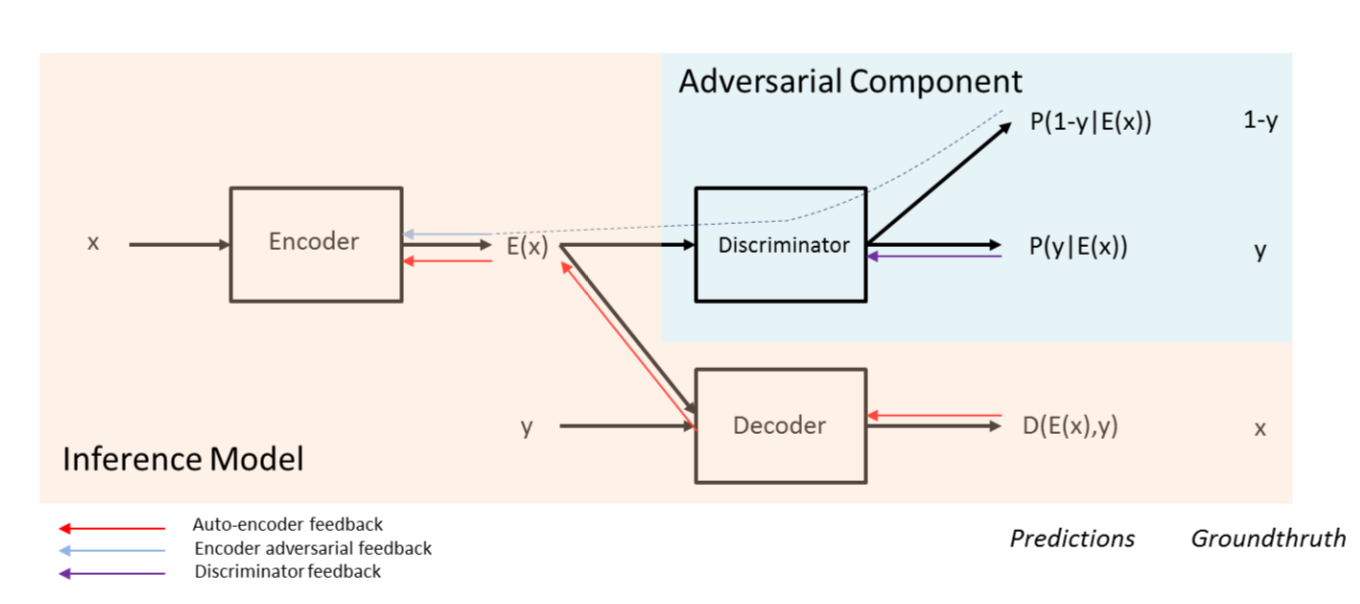}
\caption{FaderNet Architecture. Image taken from \cite{lample2017fader}}
\label{fig:1}
\end{figure}

Let $X$ be an input image and $y$ be an attribute of the image i.e. race of the person in the image. Our aim is to obtain a latent representation $X'$ which has no information of race but sufficient information of the input such that given the race value, the original image can be generated. The model is a based on an encoder-decoder architecture with adversarial training on the latent space. The architecture is drawn in Figure~\ref{fig:1}.

\paragraph{Encoder-decoder architecture} 

The encoder $E_{\theta_{enc}}: X \longrightarrow R^N$ is a convolutional neural network with parameters $\theta_{enc}$ which maps the input image to a $N$=dimensional latent space. The decoder $D_{\theta_{dec}}:(R^N, y) \longrightarrow X$ is a deconvolutional network with parameters $\theta_{dec}$ that reconstructs the images given it's latent representation and the attribute value. The loss for the encoder-decoder network (autoencoder) is simply the mean squared error of the reconstructed image and the input image:

$$ L_{ae}(\theta_{enc}, \theta_{dec}) = \frac{1}{m} \sum_{(X, y) \in T}(D(E(x), y) - X)^2 $$

where $T$ is the training set and $m$ is the number of samples in the training set.

Ideally, modifying $y$ in $D(E(x), y)$ would generate images with different perceived attributes, but similar to x in every other aspect. However, without additional constraints, the decoder learns to ignore the input $y$, and modifying $y$ at test time has no effect.

\paragraph{Learning attribute invariant representation with discriminator} 

To force the decoder to consider $y$, we need to make sure the information pertaining to attribute $y$ isn't present in $E(X)$. This is to say that given two images, $x_1$ and $x_2$, that are same except in  their attribute values should have the same latent representations $E(x_1)$ and $E(x_2)$. To generate back the original images, $x_1$ and $x_2$, the decoder is compelled to use the attribute input $y$ given to it.

This constraint is obtained by doing adversarial training on the latent space. An additional neural network, the discriminator, is trained to identify the attribute value front he latent representation. The endcoder is trained such that the discriminator is unable to identify the attribute value.  As in GANs, this corresponds to a two-player game where the discriminator aims at maximizing its ability to identify attributes, and encoder aims at preventing it to be a good discriminator.

The discriminator outputs a distribution over all possible attribute values. It's objective is given by:

$$ L_{dis}(\theta_{dis}|\theta_{enc}) = \frac{-1}{m} \sum_{(X, y) \in T} \log P_{\theta_dis}(y|E_{\theta_{enc}}(x))  $$

\paragraph{Adversarial objective} 

The objective of the encoder is now to compute a latent representation that optimizes two objectives. First, the decoder should be able to reconstruct $x$ given $E(x)$ and $y$, and at the same time the discriminator should not be able to predict $y$ given $E(x)$. Given the discriminator's parameter, the overall objective of the encoder-decoder model is:

$$ L(\theta_{enc}, \theta_{dec} |\theta_{dis}) = -\frac{1}{m} \sum_{(x, y) \in D} \\ \norm{D_{\theta_{dec}}(E_{\theta_{enc}(x, y)}) - x}^2 -
\lambda_E \log P_{\theta_{dis}}(1 - y|E_{\theta_{enc}(x)}) $$

The parameter $\lambda_E$ controls the trade-off between the reconstruction quality and the invariance of latent representation to the attributes. High values would restrain the amount of information of $x$ in $E(x)$ resulting in low quality reconstructions. Low values would limit the decoder's dependency on $y$ and thus resulting in less invariant latent representations.

One can also think of the above loss function as doing reconstruction under the constraint that the encoding of $x$ should fool the discriminator. So intuitively, using the Lagrange multipliers, we should arrive at a similar loss function and $\lambda_E$, therefore, is the constant we multiply our constraint with.
\paragraph{Training regime}

Given the overall objective function of the model, we now describe the learning algorithm. The optimal discriminator parameters satisfy $\theta^*_{dis}(\theta_{enc} \in argmin_{\theta_{dis}} L_{dis}(\theta_{dis}| \theta_{enc}))$. In practice it would be unreasonable to solve for $\theta^*_{dis}(\theta_{enc}$ for every step of encoder update. Following the practice of adversarial training for deep networks, we use stochastic gradient updates for all parameters, considering the current value of $\theta_{dis}$ as an approximation for $\theta^*_{dis}(\theta_{enc}$. Thus, the update at time $t$ given $\theta^t_{dis}$, $\theta^t_{enc}$, $\theta^t_{dec}$ and the training data point $(x_t, y_t)$ is given by,

$$ \theta^{t + 1}_{dis} = \theta^{t}_{dis} - \eta \nabla_{\theta_{dis}} L_{dis}(\theta^t_{dis} | \theta^t_{enc}, x_t, y_t) $$
$$ [\theta^{t + 1}_{enc}, \theta^{t + 1}_{dec}] = [\theta^{t}_{enc}, \theta^{t}_{dec}] - \eta \nabla_{\theta_{enc}, \theta_{dec}} L(\theta^t_{enc}, \theta^t_{dec} |\theta^{t + 1}_{dis}, x_t, y_t) $$

\paragraph{Implementation details}

The model was implemented similar to as done in \cite{lample2017fader}. The input image has the size $256 \times 256$. Let $C_k$ be a Convolution-BatchNorm-ReLU layer with $k$ filters. Convolutions use kernel of size $4 \times 4$, with a stride of $2$, and a padding of $1$, so that each layer of the encoder divides the size of its input by $2$. Leaky-ReLUs with a slope of $0.2$ in the encoder, and simple ReLUs in the decoder are used. The encoder consists of six layers as follows : 
$$ C_{16} - C_{32} - C_{64} - C_{128} - C_{256} - C_{512} $$

Thus, the latent representations have the size of $512 \times 4 \times 4$. To provide the decoder with input attributes we concatenate the input to each layer with attribute latent code i.e. one hot vectors representing the attribute values. In the UTKFace dataset we used for experiments, each image is annotated with one of 5 races. Thus the one-hot representation looks like [0, 1, 0, 0, 0]. Thus the resulting decoder has the following topology.

$$ C_{512 + 5} - C_{256 + 5} - C_{128 + 5} - C_{64 + 5} - C_{32 + 5} - C_{16 + 5} $$

The discriminator is a $C_{512}$ layer followed by a fully connected neural network of two layers with size $512$ and $5$ respectively. The model selection was done by referring to the discriminator accuracy. In principle, the accuracy of the discriminator would oscillate back and forth with an increasing trend in the beginning and then gradually falling down, ideally, to a value below 20\% considering we have five classes to classify into. The discriminator accuracy reached a highest of 94.3\% during training process. Due to computational resource limitations and also due to small dataset size, the accuracy was only reduced to 61\%. However, even with such sub-optimal accuracy, we could achieve promising results as will be shown in the following sections.

\subsection{Gender Classifiers}

We use a simple Convolution Neural Network to classify gender. The $512 \times 4 \times 4$ dimensional latent representation output from the encoder trained according to the regime described above is input to the classifier with topology as $C_{512}$ - $C_{128}$ - $MaxPool_2$ - $C_{64}$ - $C_{16}$ - $FC$.  Convolutions use kernel of size $3 \times 3$, with a stride of $2$, and a padding of $1$, so that input size remains the same after each layer of the encoder. Each convolution layer is followed by batchnorm and relu non-linearity. We also apply dropout to the first, third, fourth conv layers and the FC layer. We refer to this classifer as \textit{FaderCNN}.

We compare the above model's performance with a classifier trained with inputs that are not adversarially trained. We reduce the $256 \times 256$ sized input images to the size of latent representations obtained from FaderNet. To this end, we train a vanilla autoencoder with the exact same topology as the FaderNet, except there is no attribute latent code concatenated to input of each layer in the decoder. A CNN classifier with the exact same topology as before is trained on the latent representations obtained from this vanilla autoencoder. We refer to this classifier as \textit{SimpleCNN}.

Another baseline is the SimpleCNN with weighted loss. We refer to this model as \textit{SimpleCNN-WL}
. In this model, we weight each race category based on their inverse frequency of occurrence. Thereafter, we weight the loss from each sample with the weight of the corresponding race class and use this weighted loss function for optimizing our gender classifier.
\section{Experiments}

We evaluate our method of alleviating the bias in gender classifiers by stratifying the dataset based on the race annotation and comparing the difference in the accuracy of models on different races.

\subsection{Dataset}

\begin{figure}[h!]
\centering
\includegraphics[scale=0.5]{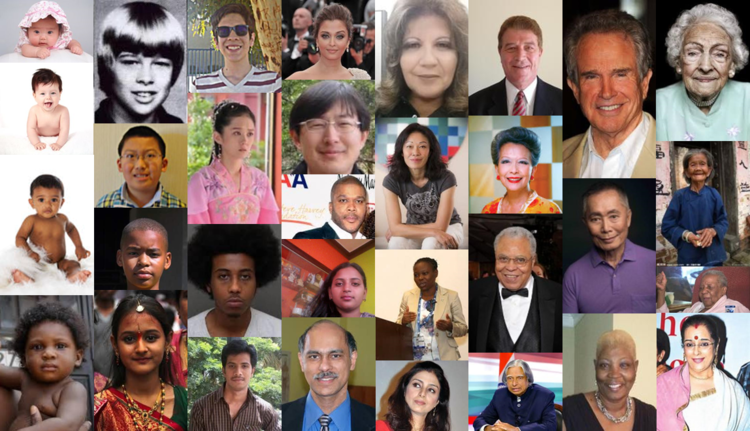}
\caption{UTKFace Examples. Image taken from \url{http://aicip.eecs.utk.edu/wiki/UTKFace}}
\label{fig:2}
\end{figure}

\begin{figure}[h!]
\centering
\includegraphics[scale=0.5]{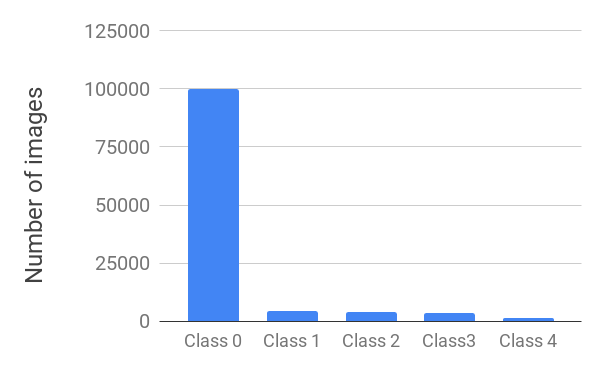}
\caption{UTKFace Dataset Statistics}
\label{fig:3}
\end{figure}

The UTKFace dataset contains images of different faces of various ethnicities and age groups. The dataset has 23,070 images. The age groups range from $0-116$. The gender of each face is labelled either male or female. The images are also labelled with five ethnicities which are White, Black, Asian, Indian and Others(like Hispanic, Latino and Middle Eastern). Some example images from the dataset is shown in Figure~\ref{fig:2}. Also, we present the dataset statistics in Figure~\ref{fig:3}.

\textbf{Preprocessing} We re-size the images to $256\times256$ from the original size of $200\times200$ and extract the attributes from the file names. There were a few wrongly labelled images which we corrected manually.

\textbf{Train, Test Split} We first create the test set by randomly selecting $474$ images from the dataset for each of the $5$ races under consideration. Thus, we get a test set of size $2370$ which is about $10\%$ of the whole dataset. We also keep another $10\%$ of the training data for validation and use the rest for training. All our evaluations have been done on the test set.


\subsection{Results}

We evaluate the three models described above on the stratified test splits. To reiterate, about 88\% of the training data are of Class 0, 4\% Class 1, 3.5\% Class 2, 3\% Class 3, and 1.5\% Class 4. At a glance, we can see from the Table \ref{results} that the accuracy on Class 2 is significantly lower than the average for the \textit{SimpleCNN} model. The sense of bias in the model can be captured by variance in the performance of the model across all classes. The \textit{SimpleCNN} model has a variance of 11.26. With the weighted loss for the same model the variance is reduced, not by a large margin, to 10.11. With our method, the variance falls down to 6.66. This proves our hypothesis that race invariant representations would indeed make our model less biased. 

\begin{table}[h!]
\centering
\vspace{2ex}
\begin{tabular}{|c|c|c|c|}
\hline
Models & \textbf{SimpleCNN} & \textbf{SimpleCNN-WL} & \textbf{FaderCNN} \\ \hline
All classes & 89.51 & 88.80 & 84.83\\ \hline
Class 0 & 92.41 & 89.84 & 85.65 \\ \hline
Class 1 & 91.56 & 90.25 & 86.08 \\ \hline
Class 2 & 84.07 & 83.64 & 80.90 \\ \hline
Class 3 & 90.93 & 92.02 & 87.66 \\ \hline
Class 4 & 88.61 & 88.27 & 83.86 \\ \hline
\textbf{Variance} & \textbf{11.26} & \textbf{10.11} & \textbf{6.66} \\ \hline

\end{tabular}
\caption{Classwise Evaluation Results on the test set}
\label{results}
\end{table}

\paragraph{Possible improvements and future work}
This work only provides as proof of concept. Note that even though the variance has reduced with both weighted loss method and our method the overall accuracy for both models has reduced, more so for FaderCNN. This can be explained by the fact that with the current training state of FaderNet, we cannot guarantee that the adversarial training has only removed the race information from the latent representation. As mentioned earlier in the implementation details, due to the nature of adversarial training and limited computation resources we could not train the model for too long and had to stop training at discriminator accuracy of 61\% even though there was a negative trend in the discriminator accuracy. We also could not extensively search the hyperparameter space for an optimal point due to computational limitations. Further, we only explored very naive data augmentation techniques like dynamic horizontal and vertical flips. Given that our dataset is very small, employing a varied set of data augmentation techniques along with some hyperparameter optimization should give much more robust results.
\section{Conclusion}

In this work, we presented a new approach to mitigate the problem of bias in simple gender classification systems. The approach is based on the recently successful methods of enforcing invariance w.r.t. the attributes by adversarial training. In particular, we remove race information from the representation of each image and try to do gender classification using this disentangled representation. We seem to have lower variance among the performance across different race classes as compared to the baseline classifiers. Though we focus on the issue of bias on race for gender classification, this approach can be employed in various other tasks with similar setting.

\bibliography{iclr2019_conference}
\bibliographystyle{iclr2019_conference}

\end{document}

%% file: math_commands.tex
\usepackage{amsmath,amsfonts,bm}

\def\eqref#1{equation~\ref{#1}}

\def\1{\bm{1}}

\DeclareMathAlphabet{\mathsfit}{\encodingdefault}{\sfdefault}{m}{sl}
\SetMathAlphabet{\mathsfit}{bold}{\encodingdefault}{\sfdefault}{bx}{n}

